\documentclass{article}

     \usepackage[final,nonatbib]{neurips_2020}

\usepackage[utf8]{inputenc} 
\usepackage[T1]{fontenc}    
\usepackage{hyperref}       
\usepackage{url}            
\usepackage{booktabs}       
\usepackage{amsfonts}       
\usepackage{amsmath}       
\usepackage{nicefrac}       
\usepackage{microtype}      
\usepackage{setspace}
\usepackage{xcolor}
\usepackage{graphicx}
\usepackage{multicol}
\usepackage{multirow}

\newcommand{\ours}{LoCo}

\definecolor{darkgreen}{rgb}{0.0, 0.7, 0.13}

\title{LoCo: Local Contrastive Representation Learning}

\author{
Yuwen Xiong\\
Uber ATG\\
University of Toronto\\
\texttt{yuwen@uber.com} \\
\And
Mengye Ren\\
Uber ATG\\
University of Toronto\\
\texttt{mren3@uber.com} \\
\And 
Raquel Urtasun\\
Uber ATG\\
University of Toronto\\
\texttt{urtasun@uber.com}
}

\begin{document}

\maketitle

\begin{abstract}
\looseness=-1
Deep neural nets typically perform end-to-end backpropagation to learn the weights, a procedure that
creates synchronization constraints in the weight update step across layers and is not biologically
plausible. Recent advances in unsupervised contrastive representation learning invite the question
of whether a learning algorithm can also be made local, that is, the updates of lower layers do not
directly depend on the computation of upper layers. While Greedy InfoMax~\cite{e2e2e} separately
learns each block with a local objective, we found that it consistently hurts readout accuracy in
state-of-the-art unsupervised contrastive learning algorithms, possibly due to the greedy objective
as well as gradient isolation. In this work, we discover that by overlapping local blocks stacking
on top of each other, we effectively increase the decoder depth and allow upper blocks to implicitly
send feedbacks to lower blocks. This simple design closes the performance gap between local learning 
and end-to-end contrastive learning algorithms for the first time. Aside from standard ImageNet 
experiments, we also show results on complex downstream tasks such as object detection and instance 
segmentation directly using readout features.

\end{abstract}
\section{Introduction}

Most deep learning algorithms nowadays are trained using backpropagation in an end-to-end fashion:
training losses are computed at the top layer and weight updates are computed based on the gradient
that flows from the very top. Such an algorithm requires lower layers to ``wait'' for upper layers,
a synchronization constraint that seems very unnatural in truly parallel distributed processing.
Indeed, there are evidences that weight synapse updates in the human brain are achieved through
local learning, without waiting for neurons in other parts of the brain to finish their jobs
\cite{caporale2008spike,ystdp}. In addition to biological plausibility aims, local learning
algorithms can also significantly reduce memory footprint during training, as they do not require
saving the intermediate activations after each local module finish its calculation. With these synchronization constraints removed, one can further
enable model parallelism in many deep network architectures \cite{pipedream} for faster parallel
training and inference.

One main objection against local learning algorithms has always been the need for supervision from
the top layer. This belief has  recently been challenged by the success of numerous self-supervised
contrastive learning algorithms~\cite{cmc,moco,pirl,simclr}, some of which can achieve matching
performance compared to supervised counterparts, meanwhile using zero class labels during the
representation learning phase. Indeed, L{\"{o}}we et al. \cite{e2e2e} show that they can separately
learn each block of layers using local contrastive learning by putting gradient stoppers in between
blocks. While the authors show matching or even sometimes superior performance using local
algorithms, we found that their gradient isolation blocks still result in degradation in accuracy in
state-of-the-art self-supervised learning frameworks, such as SimCLR~\cite{simclr}. We hypothesize
that, due to gradient isolation, lower layers are unaware of the existence of upper layers, and thus
failing to deliver the full capacity of a deep network when evaluating on large scale datasets such
as ImageNet~\cite{deng2009imagenet}.

To bridge the gradient isolation blocks and allow upper layers to influence lower layers while
maintaining localism, we propose to group two blocks into one local unit and share the middle block
simultaneously by two units. As shown in the right part of Fig.~\ref{fig:prev_model}. Thus, the
middle blocks will receive gradients from both the lower portion and the upper portion, acting like
a gradient ``bridge''. We found that such a simple scheme significantly bridges the performance gap
between Greedy InfoMax~\cite{e2e2e} and the original end-to-end algorithm~\cite{simclr}.

On ImageNet unsupervised representation learning benchmark, we evaluate  our new local learning
algorithm, named {\ours},  on both ResNet~\cite{he2016deep} and ShuffleNet~\cite{ma2018shufflenet}
architectures and found the conclusion to be the same. Aside from ImageNet object classification, we
further validate the generalizability of locally learned features on other downstream tasks such as
object detection and semantic segmentation, by only training the readout headers. On all benchmarks,
our local learning algorithm once again closely matches the more costly end-to-end trained models.

We first review related literature in local learning rules and unsupervised representation learning
in Section~\ref{sec:related}, and further elaborate the background and the two main baselines
SimCLR~\cite{simclr} and Greedy InfoMax~\cite{e2e2e} in Section~\ref{sec:gim}.
Section~\ref{sec:method} describes our {\ours} algorithm in detail. Finally, in
Section~\ref{sec:exp}, we present ImageNet-1K~\cite{deng2009imagenet} results, followed by instance
segmentation results on MS-COCO~\cite{mscoco} and Cityscapes~\cite{cityscapes}.
\section{Related Work}
\label{sec:related}

\paragraph{Neural network local learning rules:} Early neural networks literature, inspired by
biological neural networks, makes use of local associative learning rules, where the change in
synapse weights only depends on the pre- and post-activations. One classic example is the Hebbian
rule~\cite{hebb}, which strengthens the connection whenever two neurons fire together. As this can
result in numerical instability, various modifications were also proposed~\cite{oja,bcm}. These
classic learning rules can be empirically observed through long-term potentiation (LTP) and long
term depression (LTD) events during spike-timing-dependent plasticity
(STDP)~\cite{stdp,caporale2008spike}, and various computational learning models have also been
proposed~\cite{ystdp}. Local learning rules are also seen in learning algorithms such as restricted
Boltzmann machines (RBM) ~\cite{bm,hinton2012practical,dbn}, greedy layer-wise
training~\cite{greedypretrain,belilovsky2018greedy} and TargetProp~\cite{targetprop}. More recently,
it is also shown to be possible to use a network to predict the weight changes of another
network~\cite{synthetic,metaunsup,learn2remember}, as well as to learn the meta-parameters of a
plasticity rule~\cite{diffplasticity,backpropamine}. Direct feedback alignment~\cite{dfa} on the
other hand proposed to directly learn the weights from the loss to each layer by using a random
backward layer. Despite numerous attempts at bringing biological plausibility to deep neural 
networks, the performances of these learning algorithms are still far behind state-of-the-art 
networks that are trained via end-to-end backpropagation on large scale datasets. A major difference
from prior literature is that, both GIM~\cite{e2e2e} and our LoCo use an entire downsampling stage
as a unit of local computation, instead of a single convolutional layer. In fact, different
downsampling stages have been found to have rough correspondence with the primate visual
cortex~\cite{brainscore,contrastivebrain}, and therefore they can probably be viewed as better
modeling tools for local learning. Nevertheless, we do not claim to have solved the local learning
problem on a more granular level.

\paragraph{Unsupervised \& self-supervised representation learning:} Since the success of
AlexNet~\cite{alexnet}, tremendous progress has been made in terms of learning representations
without class label supervision. One of such examples is self-supervised training
objectives~\cite{selfsupbench}, such as predicting context~\cite{context,jigsaw}, predicting
rotation~\cite{rotation}, colorization~\cite{colorization} and counting~\cite{counting}.
Representations learned from these tasks can be further decoded into class labels by just training a
linear layer. Aside from predicting parts of input data, clustering objectives are also
considered~\cite{localagg,deepcluster}. Unsupervised contrastive learning has recently emerged as a
promising direction for representation learning~\cite{cpc,cmc,moco,pirl,simclr}, achieving
state-of-the-art performance on ImageNet, closing the gap between supervised training and
unsupervised training with wider networks~\cite{simclr}. Building on top of the InfoMax contrastive
learning rule~\cite{cpc}, Greedy InfoMax (GIM)~\cite{e2e2e} proposes to learn each local stage with
gradient blocks in the middle, effectively removing the backward dependency. This is similar to
block-wise greedy training~\cite{belilovsky2018greedy} but in an unsupervised fashion. 

\paragraph{Memory saving and model parallel computation:} By removing the data dependency in the
backward pass, our method can perform model parallel learning, and activations do not need to be
stored all the time to wait from the top layer. GPU memory can be saved by recomputing the
activations at the cost of longer training time~\cite{sublinear,dp,revnet}, whereas local learning
algorithms do not have such trade-off. Most parallel trainings of deep neural networks are achieved
by using data parallel training, with each GPU taking a portion of the input examples and then the
gradients are averaged. Although in the past model parallelism has also been used to vertically
split the network~\cite{alexnet,weirdtrick}, it soon went out of favor since the forward pass needs
to be synchronized. Data parallel training, on the other hand, can reach generalization bottleneck
with an extremely large batch size~\cite{dataparallel}. Recently, \cite{huang2019gpipe,pipedream}
proposed to make a pipeline design among blocks of neural networks, to allow more forward passes
while waiting for the top layers to send gradients back. However, since they use end-to-end
backpropagation, they need to save previous activations in a data buffer to avoid numerical errors
when computing the gradients. By contrast, our local learning algorithm is a natural fit for model
parallelism, without the need for extra activation storage and wait time.
\section{Background: Unsupervised Contrastive Learning}

In this section, we introduce relevant background on unsupervised contrastive learning using the
InfoNCE loss~\cite{cpc}, as well as  Greed InfoMax~\cite{e2e2e}, a local learning algorithm that
aims to learn each neural network stage with a greedy objective.

\subsection{Unsupervised Contrastive Learning \& SimCLR}

Contrastive learning~\cite{cpc} learns representations from data organized in similar or dissimilar
pairs. During  learning, an encoder is used to learn meaningful representations and a decoder is
used to distinguish the positives from the negatives through the InfoNCE loss function~\cite{cpc},
\begin{equation}
\mathcal{L}_{q, k^+, \{k^-\}} = -\log \frac{\exp(q{\cdot}k^+ / \tau)}{\exp(q{\cdot}k^+ / \tau) +
{\displaystyle\sum_{k^-}}\exp(q{\cdot}k^-  / \tau)}.
\label{eq:infonce}
\end{equation}
As shown above, the InfoNCE loss is essentially cross-entropy loss for classification with a
temperature scale factor $\tau$, where $q$ and $\{k\}$ are normalized representation vectors from
the encoder. The positive pair $(q, k^+)$ needs to be classified among all $(q, k)$ pairs. Note that
since the positive samples are defined as augmented version of the same example, this learning
objective does not need any class label information. After learning is finished, the decoder part
will be discarded and the encoder's outputs will be served as learned representations.

Recently, Chen et al. proposed SimCLR~\cite{simclr}, a state-of-the-art framework for contrastive
learning of visual representations. It proposes many useful techniques for closing the gap between
unsupervised and supervised representation learning. First, the learning benefits from a larger
batch size (\textasciitilde 2k to 8k) and stronger data augmentation. Second, it uses a non-linear
MLP projection head instead of a linear layer as the decoder, making the representation more general
as it is further away from the contrastive loss function. With 4$\times$ the channel size, it is
able to match the performance of a fully supervised ResNet-50. In this paper, we use the SimCLR
algorithm as our end-to-end baseline as it is the current state-of-the-art. We believe that our
modifications can transfer to other contrastive learning algorithms as well.

\subsection{Greedy InfoMax}
\label{sec:gim}
As unsupervised learning has achieved tremendous progress, it is natural to ask whether we can
achieve the same from a local learning algorithm. Greedy InfoMax (GIM)~\cite{e2e2e} proposed to
learn representation locally in each stage of the network, shown in the middle part of
Fig.~\ref{fig:prev_model}. It divides the encoder into several stacked modules, each with a
contrastive loss at the end. The input is forward-propagated in the usual way, but the gradients do
not propagate backward between modules. Instead, each module is trained greedily using a local
contrastive loss. This work was proposed prior to SimCLR and achieved comparable results to
CPC~\cite{cpc}, an earlier work, on a small scale dataset STL-10~\cite{coates2011analysis}. In this
paper, we used SimCLR as our main baseline, since it has superior performance on ImageNet, and we
apply the changes proposed in GIM on top of SimCLR as our local learning baseline. In our
experiments, we find that simply applying GIM on SimCLR results in a significant loss in performance
and in the next section we will explain our techniques to bridge the performance gap.
\section{LoCo: Local Contrastive Representation Learning}
\label{sec:method}
In this section, we will introduce our approach to close the gap between local contrastive learning and state-of-the-art end-to-end learning. 

\begin{figure}[t]
\vspace{-0.2in}
  \centering
  \includegraphics[width=0.98\textwidth,clip]{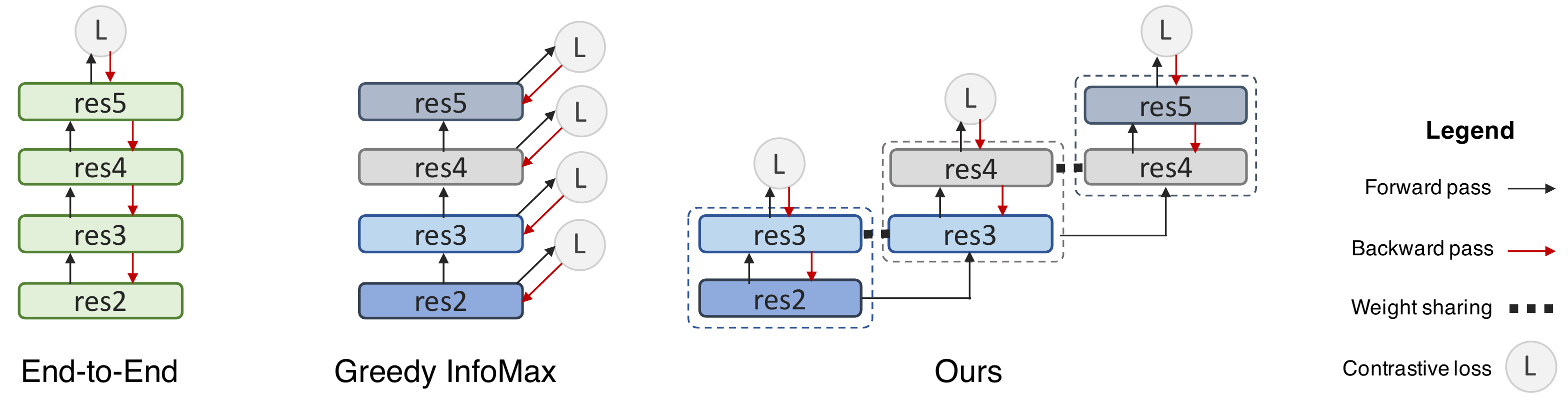}
  \caption{Comparison between End-to-End, Greedy InfoMax (GIM) and {\ours}}
  \label{fig:prev_model}
\vspace{-0.1in}
\end{figure}

In the left part of Fig.~\ref{fig:prev_model}, we show a regular end-to-end network using
backpropagation, where each rectangle denotes a downsample stage. In ResNet-50, they are {\em
conv1+res2}, {\em res3}, {\em res4}, {\em res5}. In the middle we show GIM~\cite{e2e2e}, where an
InfoNCE loss is added at the end of each local stage, and gradients do not flow back from upper
stages to lower stages. Our experimental results will show that such practice results in much worse
performance on large-scale datasets such as ImageNet. We hypothesize that it may be due to a lack of
feedback from upper layers and a lack of depth in terms of the decoders of lower layers, as they are
trying to greedily solve the classification problem. Towards fixing these two potential problems, on
the right hand side of Fig.~\ref{fig:prev_model} we show our design: we group two stages into a
unit, and each middle stage is simultaneously shared by two units. Next, we will go into details
explaining our reasonings behind these  design choices.

\subsection{Bridging the Gap between Gradient Isolation Blocks}
\label{sec:gradient_isolation}
First, in GIM, the feedback from high-level features is absent. When the difficulty of the
contrastive learning task increases (e.g., learning on a large-scale dataset such as ImageNet), the
quality of intermediate representations from lower layers will largely affect the final performance
of  upper layers. However, such demand cannot be realized because lower layers are unaware of what
kind of representations are required from above.

To overcome this issue, we hypothesize that it is essential to build a ``bridge'' between a lower
stage and its upper stage so that it can receive feedback that would otherwise be lost. As shown in
Fig.~\ref{fig:prev_model}, instead of cutting the encoder into several non-overlapping parts, we can
overlap the adjacent local stages. Each stage now essentially performs a ``look-ahead'' when
performing local gradient descent. By chaining these overlapped blocks together, it is now possible
to send feedback from the very top.

It is worth noting that, our method does not change the forward pass, even though {\em res3} and
{\em res4} appear twice in Fig.~\ref{fig:prev_model}, they receive the same inputs (from {\em res2}
and {\em res3}, respectively). Therefore the forward pass only needs to be done once in these
stages, and only the backward pass is doubled.

\subsection{Deeper Decoder}
\label{sec:deeper_decoder}
Second, we hypothesize that the receptive field of early stages in the encoder might be too small to
effectively solve the contrastive learning problem. As the same InfoNCE function is applied to all
local learning blocks (both early and late stages), it is difficult for the decoder to use
intermediate representation from the early stages to successful classify the positive sample,
because of the limitation of their receptive fields. For example, in the first stage, we need to
perform a global average pooling on the entire feature map with a spatial dimension of $56\times56$
before we send it to the decoder for classification.

In Section~\ref{sec:exp}, we empirically verify our hypothesis by showing that adding convolutional
layers into the decoder to enlarge the receptive field is essential for local algorithms. However,
this change does not show any difference in the end-to-end version with a single loss, since the
receptive field of the final stage is already large enough. Importantly, by having an overlapped
stage shared between local units, we effectively make decoders deeper without introducing extra cost
in the forward pass, simultaneously solving both issues described in this section.
\section{Experiments}
\label{sec:exp}

In this section, we conduct experiments to test the  hypotheses we made in Section~\ref{sec:method}
and verify our design choices. Following previous works~\cite{colorization, cpc,
bachman2019learning, kolesnikov2019revisiting, moco}, we first evaluate the quality of the learned
representation using ImageNet~\cite{deng2009imagenet}, followed by results on MS-COCO~\cite{mscoco}
and Cityscapes~\cite{cityscapes}. We use SimCLR~\cite{simclr} and GIM~\cite{e2e2e} as our main
baselines, and  consider both ResNet-50~\cite{he2016deep} and ShuffleNet
v2-50~\cite{ma2018shufflenet} backbone architectures as the encoder network.

\subsection{ImageNet-1K} 
\paragraph{Implementation details:} Unless otherwise specified, we train with a batch size of 4096
using the LARS optimizer~\cite{you2017large}. We train models 800 epochs to show that \ours{} can
perform well on very long training schedules and match state-of-the-art performance; we use a
learning rate of 4.8 with a cosine decay schedule without restart~\cite{loshchilov2016sgdr}; linear
warm-up is used for the first 10 epochs. Standard data augmentations such as random cropping, random
color distortion, and random Gaussian blurring are used. For local learning algorithms (i.e., GIM 
and {\ours}), 2-layer MLPs with global average pooling are used to project the intermediate features
into a 128-dim latent space, unless otherwise specified in ablation studies.
Following~\cite{colorization, cpc, bachman2019learning, kolesnikov2019revisiting, moco}, we evaluate
the quality of the learned representation by freezing the encoder and training a linear classifier
on top of the trained encoders. SGD without momentum is used as the optimizer for 100 training
epochs with a learning rate of 30 and decayed by a factor of 10 at epoch 30, 60 and 90, the same
procedure done in~\cite{moco}.

\paragraph{Main results:} As shown in Table~\ref{tab:main_results}, SimCLR achieves favorable
results compared to other previous contrastive learning methods. For instance, CPC~\cite{cpc}, the
contrastive learning algorithm which Greedy InfoMax (GIM) was originally based on, performs much
worse. By applying GIM on top of SimCLR, we see a significant drop of 5\% on the top 1 accuracy. Our
method clearly outperforms GIM by a large margin, and is even slightly better than the end-to-end
SimCLR baseline, possibly caused by the fact that better representations are obtained via multiple
training losses applied at different local decoders.

\begin{table}
\vspace{-0.4in}
\begin{minipage}[t]{0.42\linewidth}
\resizebox{1\linewidth}{!}{
\centering
\begin{tabular}{lccc}
     \toprule
     Method                         & Architecture 		& Acc. 	&  Local           \\
     \midrule     
     Local Agg. ~\cite{localagg}    & ResNet-50			& 60.2         &             \\
     MoCo~\cite{moco}               & ResNet-50			& 60.6         &             \\
	PIRL~\cite{pirl}               & ResNet-50			& 63.6         &             \\
     CPC v2~\cite{cpc}              & ResNet-50			& 63.8         &             \\
     SimCLR*~\cite{simclr}			& ResNet-50			& 69.3		   &			 \\
     \midrule
     SimCLR~\cite{simclr}           & ResNet-50 			& {\bf 69.8}         &             \\ 
     GIM~\cite{e2e2e}	           & ResNet-50 			& 64.7         & \checkmark  \\
     \ours{} (Ours)                 & ResNet-50 			& 69.5         & \checkmark  \\
     \midrule
     SimCLR~\cite{simclr} 		 & ShuffleNet v2-50 	& 69.1         &             \\
     GIM~\cite{e2e2e}			 & ShuffleNet v2-50 	& 63.5         & \checkmark  \\
     \ours{} (Ours)                 & ShuffleNet v2-50 	& {\bf 69.3}         & \checkmark  \\
     \bottomrule
\end{tabular}	
}
\vspace{0.1in}
\caption{ImageNet accuracies of linear classifiers trained on representations learned with different
unsupervised methods, SimCLR* is the result from the SimCLR paper with 1000 training epochs. }
\label{tab:main_results}
\end{minipage}
\hfill
\begin{minipage}[t]{0.55\linewidth}
\resizebox{1\linewidth}{!}{
\centering
\begin{tabular}{lc|ll|ll}
     \toprule
     \multirow{2}*{Method} & \multirow{2}*{Arch} & \multicolumn{2}{c|}{COCO} & \multicolumn{2}{c}{Cityscapes}\\
     ~ & ~ & \multicolumn{1}{c}{AP$^\text{bb}$} & \multicolumn{1}{c|}{AP} & \multicolumn{1}{c}{AP$^\text{bb}$} & \multicolumn{1}{c}{AP}  \\
     \midrule
     Supervised		& R-50			& 33.9  	& 31.3	& 33.2 & 27.1\\
     \midrule
     \multicolumn{6}{c}{Backbone weights with 100 Epochs}\\
     \midrule
     SimCLR 		& R-50 			& 32.2  	& 29.9 	& 33.2 & 28.6\\
     GIM            & R-50 			& 27.7 \color{red}{(-4.5)}	& 25.7 \color{red}{(-4.2)}	& 30.0 \color{red}{(-3.2)} & 24.6 \color{red}{(-4.0)}\\
     Ours     		& R-50 			& 32.6 \color{black}{(+0.4)}	& 30.1 \color{black}{(+0.2)}	& 33.2 \color{black}{(+0.0)} & 28.4 \color{black}{(-0.2)}\\
     \midrule
     SimCLR 		& Sh-50 	& 32.5 	& 30.1	& 33.3 & 28.0\\
     GIM            & Sh-50 	& 27.3 \color{red}{(-5.2)}  	& 25.4 \color{red}{(-4.7)}	& 29.1 \color{red}{(-4.2)} & 23.9 \color{red}{(-4.1)}\\
     Ours     		& Sh-50 	& 31.8 \color{black}{(-0.7)}  	& 29.4 \color{black}{(-0.7)} 	& 33.1 \color{black}{(-0.2)} & 27.7 \color{black}{(-0.3)}\\
     \midrule
     \multicolumn{6}{c}{Backbone weights with 800 Epochs}\\
     \midrule
     SimCLR 		& R-50 			& 34.8  	& 32.2 	& 34.8 & 30.1 \\
     GIM            & R-50 			& 29.3 \color{red}{(-5.5)}	& 27.0 \color{red}{(-5.2)}	& 30.7 \color{red}{(-4.1)} & 26.0 \color{red}{(-4.1)} \\
     Ours     		& R-50 			& 34.5 \color{black}{(-0.3)}	& 32.0 \color{black}{(-0.2)}	& 34.2 \color{black}{(-0.6)} & 29.5 \color{black}{(-0.6)} \\
     \midrule
     SimCLR 		& Sh-50 	& 33.4 	& 30.9	& 33.9 & 28.7 \\
     GIM            & Sh-50 	& 28.9 \color{red}{(-4.5)}  	& 26.9 \color{red}{(-4.0)}	& 29.6 \color{red}{(-4.3)} & 23.9 \color{red}{(-4.8)} \\
     Ours     		& Sh-50 	& 33.6 \color{black}{(+0.2)} 	& 31.2 \color{black}{(+0.3)}	& 33.0 \color{black}{(-0.9)} & 28.1 \color{black}{(-0.6)} \\ \bottomrule
\end{tabular}	
}
\caption{Mask R-CNN results on COCO and Cityscapes. Backbone networks are frozen. ``R-50'' denotes
ResNet-50 and ``Sh-50'' denotes ShuffleNet v2-50.}
\label{tab:det_results}	
\end{minipage}
\vspace{-0.2in}
\end{table}

\subsection{Performance on Downstream Tasks}
In order to further verify the quality and generalizability of the learned representations, we use
the trained encoder from previous section as pre-trained models to perform downstream tasks, We use
Mask R-CNN~\cite{maskrcnn} on Cityscapes~\cite{cityscapes} and COCO~\cite{mscoco} to evaluate object
detection and instance segmentation performance. Unlike what has been done in MoCo~\cite{moco},
where the whole network is finetuned on downstream task, here we freeze the pretrained backbone
network, so that we better distinguish the differences in quality of different unsupervised learning
methods.

\paragraph{Implementation details:}  To mitigate the distribution gap between features from the
supervised pre-training model and contrastive learning model, and reuse the same hyperparameters
that are selected for the supervised pre-training model~\cite{moco}, we add
SyncBN~\cite{peng2018megdet} after all newly added layers in FPN and bbox/mask heads. The two-layer
MLP box head is replaced with a {\em 4conv-1fc} box head to better leverage
SyncBN~\cite{wu2018group}. We conduct the downstream task experiments using
mmdetection~\cite{mmdetection}. Following~\cite{moco}, we use the same hyperparameters as the
ImageNet supervised counterpart for all experiments, with $1\times$ ($\sim$12 epochs) schedule for 
COCO and 64 epochs for Cityscapes, respectively. Besides SimCLR and GIM, we provide one more 
baseline using weights pretrained on ImageNet via supervised learning provided by
PyTorch\footnote{\url{https://download.pytorch.org/models/resnet50-19c8e357.pth}} for reference.

\paragraph{Results:} From the Table~\ref{tab:det_results} we can clearly see that the conclusion is
consistent on downstream tasks. Better accuracy on ImageNet linear evaluation also translates to
better instance segmentation quality on both COCO and Cityscapes. {\ours} not only closes the gap
with end-to-end baselines on object classification in the training domain but also on downstream
tasks in new domains.

\looseness=-1
Surprisingly, even though SimCLR and \ours{} cannot exactly match ``Supervised'' on ImageNet, they
are 1 -- 2 points AP better than ``Supervised'' on downstream tasks. This shows unsupervised
representation learning can learn more generalizable features that are more transferable to new
domains.

\begin{table}
\vspace{-0.4in}
\begin{minipage}[t]{1.0\linewidth}
\centering
\resizebox{0.6\linewidth}{!}{
\begin{tabular}{l|ll|ll}
     \toprule
     \multicolumn{1}{c|}{Pretrain} & \multicolumn{2}{c|}{COCO-10K} & \multicolumn{2}{c}{COCO-1K} \\
     \multicolumn{1}{c|}{Method} & \multicolumn{1}{c}{AP$^\text{bb}$} & \multicolumn{1}{c|}{AP} & \multicolumn{1}{c}{AP$^\text{bb}$} & \multicolumn{1}{c}{AP} \\
     \midrule
     Random Init	& 23.5 & 22.0 & 2.5 & 2.5 \\
     Supervised		& 26.0 & 23.8 & 10.4 & 10.1 \\
     \midrule
     \multicolumn{5}{c}{Pretrained weights with 100 Epochs}\\
     \midrule
     SimCLR 		& 25.6 & 23.9 & 11.3 & 11.4 \\
     GIM 			& 22.6  \color{red}{(-3.0)} & 20.8  \color{red}{(-3.1)} & \ \  9.7  \color{red}{(-1.6)} & \ \  9.6  \color{red}{(-1.8)} \\
     Ours     		& 26.1  \color{black}{(+0.3)} & 24.2  \color{black}{(+0.5)} & 11.7  \color{black}{(+0.4)}& 11.8  \color{black}{(+0.4)} \\
     \midrule
     \multicolumn{5}{c}{Pretrained weights with 800 Epochs}\\
     \midrule
     SimCLR 		& 27.2 & 25.2 & 13.9 & 14.1 \\
     GIM 			& 24.4 \color{red}{(-2.8)} & 22.4 \color{red}{(-2.8)} & 11.5 \color{red}{(-2.4)} & 11.7 \color{red}{(-2.4)}\\
     Ours     		& 27.8 \color{black}{(+0.6)} & 25.6 \color{black}{(+0.4)} & 13.9 \color{black}{(+0.0)} & 13.8 \color{black}{(-0.3)} \\
     \bottomrule
\end{tabular}
}
\vspace{0.1in}
\caption{Mask R-CNN results on 10K COCO images and 1K COCO images}
\label{tab:semi_det_results}
\vspace{-0.3in}
\end{minipage}
\end{table}

\subsection{Downstream Tasks with Limited Labeled Data} 
With the power of unsupervised representation learning, one can learn a deep model with much less
amount of labeled data on downstream tasks. Following~\cite{he2019rethinking}, we randomly sample
10k and 1k  COCO images for training, namely COCO-10K and COCO-1K. These are 10\% and 1\% of the
full COCO train2017 set. We report AP on the official val2017 set. Besides SimCLR and GIM, we also
provide two baselines for reference: ``Supervised'' as mentioned in previous subsection, and
``Random Init'' does not use any pretrained weight but just uses random initialization for all
layers and trains from scratch.

Hyperparameters are kept the same as ~\cite{he2019rethinking} with multi-scale training except for
adjusted learning rate and decay schedules. We train models for 60k iterations (96 epochs) on
COCO-10K and 15k iterations (240 epochs) on COCO-1K with a batch size of 16. All models use
ResNet-50 as the backbone and are finetuned with SyncBN~\cite{peng2018megdet}, {\em conv1} and {\em
res2} are frozen except ``Random Init" entry. We make 5 random splits for both COCO-10K/1K and run
all entries on these 5 splits and take the average. The results are very stable and the variance is
very small ($<0.2$).

\paragraph{Results:} 
Experimental results are shown in Table~\ref{tab:semi_det_results}. Random initialization is
significantly worse than other models that are pretrained on ImageNet, in agreement with the results
reported by~\cite{he2019rethinking}. With weights pretrained for 100 epochs, both SimCLR and \ours{}
get sometimes better performance compared to supervised pre-training, especially toward the regime
of limited labels (i.e., COCO-1K). This shows that the unsupervised features are more general as
they do not aim to solve the ImageNet classification problem. Again, GIM does not perform well and
cannot match the randomly initialized baseline. Since we do not finetune early stages, this suggests
that GIM does not learn generalizable features in its early stages. We conclude that our proposed
{\ours} algorithm is able to learn generalizable features for downstream tasks, and is especially
beneficial when limited labeled data are available.

Similar to the previous subsection, we run pretraining longer until 800 epochs, and observe
noticeable improvements on both tasks and datasets. This results seem different from the one
reported in~\cite{chen2020improved} that longer iterations help improve the ImageNet accuracy but do
not improve downstream VOC detection performance. Using 800 epoch pretraining, both \ours{} and
SimCLR can outperform the supervised baseline by 2 points AP on COCO-10K and 4 points AP on COCO-1K.

\subsection{Influence of the Decoder Depth}
In this section, we study the influence of the decoder depth. First, we investigate the
effectiveness of the convolutional layers we add in the decoder. The results are shown in
Table~\ref{tab:ablation_conv_blocks}. As we can see from the ``1 conv block without local and
sharing property'' entry in the table, adding one more residual convolution block at the end of the
encoder, i.e. the beginning of the decoder, in the original SimCLR does not help. One possible
reason is that the receptive field is large enough at the very end of the encoder. However, adding
one convolution block with downsampling before the global average pooling operation in the decoder
will significantly improve the performance of local contrastive learning. We argue that such a
convolution block will enlarge the receptive field as well as the capacity of the local decoders and
lead to better representation learning even with gradient isolation. If the added convolution block
has no downsampling factor (denoted as ``w/o ds''), the improvement is not be as significant.
 
We also try adding more convolution layers in the decoder, including adding two convolution
blocks (denoted as ``2 conv blocks''), adding one stage to make the decoder as deep as the
next residual stage of the encoder (denoted as ``one stage''), as well as adding layers to make each
decoder as deep as the full Res-50 encoder (denoted as ``full network''). The results of these
entries show that adding more convolution layers helps, but the improvement will eventually diminish
and these entries achieve the same performance as SimCLR.

Lastly, we show that by adding two more layers in the MLP decoders, i.e. four layers in total, we
can observe the same amount of performance boost on all of methods, as shown in the 4th to 6th row
of Table~\ref{tab:ablation_conv_blocks}. However, increasing MLP decoder depth cannot help us bridge
the gap between local and end-to-end contrastive learning.

To reduce the overhead we introduce in the decoder, we decide to add one residual convolution block
only and keep the MLP depth to 2, as was done the original SimCLR. It is also worth noting that by
sharing one stage of the encoder, our method can already closely match SimCLR without deeper
decoders, as shown in the third row of Table~\ref{tab:ablation_conv_blocks}.

\begin{table}
\begin{minipage}[t]{0.5\linewidth}
\centering

\resizebox{1.0\linewidth}{!}{
\begin{tabular}{l|cc|c}
     \toprule
\begin{tabular}[l]{@{}l@{}} Extra Layers \\
     before MLP Decoder 	\end{tabular} & \multirow{1}*{Local} & \multirow{1}*{Sharing} 	& \multirow{1}*{Acc.} \\
     \midrule
     None		 			& & 	& 65.7 			                                                 \\
	 None 					& \checkmark	 &	& 60.9  	                                        \\
	 \midrule
     1 conv block			& & 	& 65.6 			                                                 \\
	 1 conv block (w/o ds) 	& \checkmark &     & 63.6                \\
	 1 conv block		    & \checkmark &	     & 65.1              \\
	 \midrule
	 2 conv blocks		    	& \checkmark &     & 65.8  	           \\
	 1 stage   				& \checkmark & & 65.8  	  	      \\
	 full network			& \checkmark &     & 65.8  	  	  \\
	 \midrule
     2-layer MLP		 	& & 	& 67.1 			                                            \\
	 2-layer MLP			& \checkmark & 	& 62.3   			                   \\
	 \midrule
	 Ours 	     & \checkmark 	      & \checkmark   	    & 66.2             \\
	 Ours + 2-layer MLP & \checkmark & \checkmark & {\bf 67.5} \\
     \bottomrule
\end{tabular}	
}

\vspace{0.1in}
\caption{ImageNet accuracies of models with different decoder architecture. All entries are trained
with 100 epochs. }
\label{tab:ablation_conv_blocks}
\end{minipage}
\hfill
\begin{minipage}[t]{0.45\linewidth}
\centering
\resizebox{0.7\linewidth}{!}{
\begin{tabular}{l|c}
     \toprule
     Sharing description		& Acc.       \\
     \midrule
	 No sharing			& 65.1 	  \\
	 Upper layer grad only   &  65.3     \\
     \midrule
     L2 penalty (1e-4)		& 65.5      \\
     L2 penalty (1e-3)		& 66.0      \\
     L2 penalty (1e-2)		& 65.9      \\
     \midrule
     Sharing 1 block		& 64.8      \\
     Sharing 2 blocks		& 65.3      \\
     Sharing 1 stage		& {\bf 66.2}\\     
     \bottomrule
\end{tabular}	
}
\vspace{0.1in}
\caption{ImageNet accuracies of models with different sharing strategies. All entries are trained
with 100 epochs.}
\label{tab:ablation_sharing}
\end{minipage}
\vspace{-0.2in}
\end{table}

\subsection{Influence of the Sharing Strategy} 
As we argued in Sec.~\ref{sec:gradient_isolation} that local contrastive learning may suffer from
gradient isolation, it is important to verify this situation and know how to build a feedback
mechanism properly. In Table~\ref{tab:ablation_sharing}, we explore several sharing strategies to
show their impact of the performance. All entries are equipped with 1 residual convolution block +
2-layer MLP decoders.

We would like to study what kind of sharing can build implicit feedback. In {\ours} the shared stage
between two local learning modules is updated by gradients associated with losses from both lower
and upper local learning modules. Can implicit feedback be achieved by another way? To answer this
question, we try to discard part of the gradients of a block shared in both local and upper local
learning modules. Only the gradients calculated from the loss associated with the upper module will
be kept to update the weights. This control is denoted as ``Upper layer grad only'' in
Table~\ref{tab:ablation_sharing} and the result indicates that although the performance is slightly
improved compared to not sharing any encoder blocks, it is worse than taking gradients from both
sides.

We also investigate soft sharing, i.e. weights are not directly shared in different local learning
modules but are instead softly tied using L2 penalty on the differences. For each layer in the
shared stage, e.g., layers in {\em res3}, the weights are identical in different local learning
modules upon initialization, and they will diverge as the training progress goes on. We add an L2
penalty on the difference of the weights in each pair of local learning modules, similar to L2
regularization on weights during neural network training. We try three different coefficients from
1e-2 to 1e-4 to control the strength of soft sharing. The results in
Table~\ref{tab:ablation_sharing} show that soft sharing also brings improvements but it is slightly
worse than hard sharing.  Note that with this strategy the forward computation cannot be shared and
the computation cost is increased. Thus we believe that soft sharing is not an ideal way to achieve
good performance.

Finally, we test whether sharing can be done with fewer residual convolution blocks between local
learning modules rather than a whole stage, in other words, we vary the size of the local learning
modules to observe any differences. We try to make each module contain only one stage plus a few
residual blocks at the beginning of the next stage instead of two entire stages. Therefore, only the
blocks at the beginning of stages are shared between different modules. This can be seen as a smooth
transition between GIM and {\ours}. We try only sharing the first block or first two blocks of each
stage, leading to ``Sharing 1 block'' and ``Sharing 2 blocks'' entries in
Table~\ref{tab:ablation_sharing}. The results show that sharing fewer blocks of each stage will not
improve performance and sharing only 1 block will even hurt.

\subsection{Memory Saving}

Although local learning saves GPU memory, we find that the original ResNet-50 architecture prevents
{\ours} to further benefit from local learning, since ResNet-50 was designed with balanced
computation cost at each stage and memory footprint was not taken into consideration. In ResNet,
when performing downsampling operations at the beginning of each stage, the spatial dimension is
reduced by $1/4$ but the number of channels only doubles, therefore the memory usage of the lower
stage will be twice as much as the upper stage. Such design choice makes {\em conv1} and {\em res2}
almost occupy 50\% of the network memory footprint. When using ResNet-50, the memory saving ratio
of GIM is $1.81\times$ compared to the original, where the memory saving ratio is defined as the
reciprocal of peak memory usage between two models. {\ours} can achieve $1.28\times$ memory saving
ratio since it needs to store one extra stage.

We also show that by properly designing the network architecture, we can make training benefit more
from local learning. We change the 4-stage ResNet to a 6-stage variant with a more progressive
downsampling mechanism. In particular, each stage has 3 residual blocks, leading to a Progressive
ResNet-50 (PResNet-50). Table~\ref{tbl:structures} compares memory footprint and computation of each
stage for PResNet-56 and ResNet-50 in detail. The number of base channels for each stage are 56, 96,
144, 256, 512, 1024, respectively. After {\em conv1} and {\em pool1}, we gradually downsample the
feature map resolution from 56x56 to 36x36, 24x24, 16x16, 12x12, 8x8 at each stage with bilinear
interpolation instead of strided convolution~\cite{he2016deep}. Grouped convolution~\cite{alexnet}
with 2, 16, 128 groups is used in the last three stages respectively to reduce the computation cost.
The difference between PResNet-56 and ResNet-50 and block structures are illustrated in
 appendix.

By simply making this modification without other new techniques~\cite{he2019bag, hu2018squeeze,
li2019selective}, we can get a network that matches the ResNet-50 performance with similar
computation costs. More importantly, it has balanced memory footprint at each stage. As shown in
Table~\ref{tab:memory_saving}, SimCLR using PResNet-50 gets 66.8\% accuracy, slightly better
compared to the ResNet-50 encoder. Using PResNet-50, our method performs on par with SimCLR while
still achieving remarkable memory savings of 2.76 $\times$. By contrast, GIM now has an even larger
gap (14 points behind SimCLR) compared to before with ResNet-50, possibly due to the receptive field
issue we mentioned in Sec.~\ref{sec:deeper_decoder}.

\begin{table}
\begin{minipage}[htbp]{0.5\linewidth}
\centering
\resizebox{0.9\linewidth}{!}{
\begin{tabular}{c|c|c|c|c}
\bottomrule
\multirow{2}{*}{Stage}     &  \multicolumn{2}{c|}{PResNet-50} & \multicolumn{2}{c}{ResNet-50}\\ 
\cline{2-5} & \begin{tabular}[c]{@{}c@{}}Mem. \\ (\%)\end{tabular}  & \begin{tabular}[c]{@{}c@{}}FLOPS \\ (\%)\end{tabular}& \begin{tabular}[c]{@{}c@{}}Mem. \\ (\%)\end{tabular} & \begin{tabular}[c]{@{}c@{}}FLOPS \\ (\%)\end{tabular} \\
\hline
res2& 15.46 & 13.50 & 43.64 & 19.39 \\
res3 & 10.96 & 14.63 & 29.09 & 25.09\\
res4 & 19.48 & 14.77 & 21.82 & 35.80\\
res5 & 17.31 & 16.62 & 5.45 & 19.73\\
res6 & 19.48 & 20.45 & - & - \\
res7 & 17.31 & 20.04 & - & - \\
\hline
FLOPs & \multicolumn{2}{c|}{4.16G} & \multicolumn{2}{c}{4.14G}\\
\toprule
\end{tabular}
}
\vspace{0.1in}
\caption{Memory footprint and computation percentages for PResNet-50 and ResNet-50 on stage level.}
\label{tbl:structures}
\end{minipage}
\hfill
\begin{minipage}[t]{0.45\linewidth}
\vspace{-0.8in}
\centering
\resizebox{0.7\linewidth}{!}{
\begin{tabular}{l|c|c}
     \toprule
     Method		& Acc. & \begin{tabular}[c]{@{}c@{}}Memory \\ Saving Ratio\end{tabular}  \\
     \midrule
	 SimCLR				& 66.8     & $1\times$ 			\\
	 GIM &  52.6     & $4.56\times$\\
	 \ours &  66.6     & $2.76\times$\\
     \bottomrule
\end{tabular}	
}
\vspace{0.1in}
\caption{ImageNet accuracies and memory saving ratio of Progressive ResNet-50 with balanced memory
footprint at each stage. All entries are trained with 100 epochs.}
\label{tab:memory_saving}
\end{minipage}
\vspace{-0.2in}
\end{table}
\section{Conclusion}
We have presented {\ours}, a local learning algorithm for unsupervised contrastive learning. We show
that by introducing implicit gradient feedback between the gradient isolation blocks and properly
deepening the decoders, we can largely close the gap between local contrastive learning and
state-of-the-art end-to-end contrastive learning frameworks. Experiments on ImageNet and downstream
tasks show that {\ours} can learn good visual representations for both object recognition and
instance segmentation just like end-to-end approaches can. Meanwhile, it can benefit from nice
properties of local learning, such as lower peak memory footprint and faster model parallel
training.

\section*{Broader Impact}
\vspace{-0.1in}
Our work aims to make deep unsupervised representation learning more biologically plausible by
removing the reliance on end-to-end backpropagation, a step towards a better understanding of the
learning in our brain. This can potentially lead to solutions towards mental and psychological
illness. Our algorithm also lowers the GPU memory requirements and can be deployed with model
parallel configurations. This can potentially allow deep learning training to run on cheaper and
more energy efficient hardware, which would make a positive impact to combat climate change. We
acknowledge unknown risks can be brought by the development of AI technology; however, the
contribution of this paper has no greater risk than any other generic deep learning paper that
studies standard datasets such as ImageNet.


{
\bibliography{ref}

\begin{thebibliography}{10}

\bibitem{stdp}
L.~F. Abbott and S.~B. Nelson.
\newblock Synaptic plasticity: taming the beast.
\newblock {\em Nature neuroscience}, 3(11):1178--1183, 2000.

\bibitem{bachman2019learning}
P.~Bachman, R.~D. Hjelm, and W.~Buchwalter.
\newblock Learning representations by maximizing mutual information across
  views.
\newblock In {\em Advances in Neural Information Processing Systems,
  {NeurIPS}}, 2019.

\bibitem{belilovsky2018greedy}
E.~Belilovsky, M.~Eickenberg, and E.~Oyallon.
\newblock Greedy layerwise learning can scale to imagenet.
\newblock {\em arXiv preprint arXiv:1812.11446}, 2018.

\bibitem{targetprop}
Y.~Bengio.
\newblock How auto-encoders could provide credit assignment in deep networks
  via target propagation.
\newblock {\em CoRR}, abs/1407.7906, 2014.

\bibitem{greedypretrain}
Y.~Bengio, P.~Lamblin, D.~Popovici, and H.~Larochelle.
\newblock Greedy layer-wise training of deep networks.
\newblock In B.~Sch{\"{o}}lkopf, J.~C. Platt, and T.~Hofmann, editors, {\em
  Advances in Neural Information Processing Systems, {NIPS}}, 2006.

\bibitem{ystdp}
Y.~Bengio, D.~Lee, J.~Bornschein, and Z.~Lin.
\newblock Towards biologically plausible deep learning.
\newblock {\em CoRR}, abs/1502.04156, 2015.

\bibitem{bcm}
E.~L. Bienenstock, L.~N. Cooper, and P.~W. Munro.
\newblock Theory for the development of neuron selectivity: orientation
  specificity and binocular interaction in visual cortex.
\newblock {\em Journal of Neuroscience}, 2(1):32--48, 1982.

\bibitem{caporale2008spike}
N.~Caporale and Y.~Dan.
\newblock Spike timing--dependent plasticity: a hebbian learning rule.
\newblock {\em Annu. Rev. Neurosci.}, 31:25--46, 2008.

\bibitem{deepcluster}
M.~Caron, P.~Bojanowski, A.~Joulin, and M.~Douze.
\newblock Deep clustering for unsupervised learning of visual features.
\newblock In V.~Ferrari, M.~Hebert, C.~Sminchisescu, and Y.~Weiss, editors,
  {\em 15th European Conference on Computer Vision, {ECCV}}, 2018.

\bibitem{mmdetection}
K.~Chen, J.~Wang, J.~Pang, Y.~Cao, Y.~Xiong, X.~Li, S.~Sun, W.~Feng, Z.~Liu,
  J.~Xu, Z.~Zhang, D.~Cheng, C.~Zhu, T.~Cheng, Q.~Zhao, B.~Li, X.~Lu, R.~Zhu,
  Y.~Wu, J.~Dai, J.~Wang, J.~Shi, W.~Ouyang, C.~C. Loy, and D.~Lin.
\newblock {MMDetection}: Open mmlab detection toolbox and benchmark.
\newblock {\em arXiv preprint arXiv:1906.07155}, 2019.

\bibitem{simclr}
T.~Chen, S.~Kornblith, M.~Norouzi, and G.~E. Hinton.
\newblock A simple framework for contrastive learning of visual
  representations.
\newblock {\em CoRR}, abs/2002.05709, 2020.

\bibitem{sublinear}
T.~Chen, B.~Xu, C.~Zhang, and C.~Guestrin.
\newblock Training deep nets with sublinear memory cost.
\newblock {\em CoRR}, abs/1604.06174, 2016.

\bibitem{chen2020improved}
X.~Chen, H.~Fan, R.~Girshick, and K.~He.
\newblock Improved baselines with momentum contrastive learning.
\newblock {\em arXiv preprint arXiv:2003.04297}, 2020.

\bibitem{coates2011analysis}
A.~Coates, A.~Ng, and H.~Lee.
\newblock An analysis of single-layer networks in unsupervised feature
  learning.
\newblock In {\em 14th International Conference on Artificial Intelligence and
  Statistics, {AISTATS}}, 2011.

\bibitem{cityscapes}
M.~Cordts, M.~Omran, S.~Ramos, T.~Rehfeld, M.~Enzweiler, R.~Benenson,
  U.~Franke, S.~Roth, and B.~Schiele.
\newblock The cityscapes dataset for semantic urban scene understanding.
\newblock In {\em {IEEE} Conference on Computer Vision and Pattern Recognition,
  {CVPR}}, 2016.

\bibitem{deng2009imagenet}
J.~Deng, W.~Dong, R.~Socher, L.-J. Li, K.~Li, and L.~Fei-Fei.
\newblock Imagenet: A large-scale hierarchical image database.
\newblock In {\em IEEE Conference on Computer Vision and Pattern Recognition,
  {CVPR}}, 2009.

\bibitem{context}
C.~Doersch, A.~Gupta, and A.~A. Efros.
\newblock Unsupervised visual representation learning by context prediction.
\newblock In {\em {IEEE} International Conference on Computer Vision, {ICCV}},
  2015.

\bibitem{rotation}
S.~Gidaris, P.~Singh, and N.~Komodakis.
\newblock Unsupervised representation learning by predicting image rotations.
\newblock In {\em 6th International Conference on Learning Representations,
  {ICLR}}, 2018.

\bibitem{revnet}
A.~N. Gomez, M.~Ren, R.~Urtasun, and R.~B. Grosse.
\newblock The reversible residual network: Backpropagation without storing
  activations.
\newblock In {\em Advances in Neural Information Processing Systems, {NIPS}},
  2017.

\bibitem{selfsupbench}
P.~Goyal, D.~Mahajan, A.~Gupta, and I.~Misra.
\newblock Scaling and benchmarking self-supervised visual representation
  learning.
\newblock In {\em 2019 {IEEE/CVF} International Conference on Computer Vision,
  {ICCV}}, 2019.

\bibitem{dp}
A.~Gruslys, R.~Munos, I.~Danihelka, M.~Lanctot, and A.~Graves.
\newblock Memory-efficient backpropagation through time.
\newblock In {\em Advances in Neural Information Processing Systems, {NIPS}},
  2016.

\bibitem{moco}
K.~He, H.~Fan, Y.~Wu, S.~Xie, and R.~B. Girshick.
\newblock Momentum contrast for unsupervised visual representation learning.
\newblock {\em CoRR}, abs/1911.05722, 2019.

\bibitem{he2019rethinking}
K.~He, R.~Girshick, and P.~Doll{\'a}r.
\newblock Rethinking imagenet pre-training.
\newblock In {\em IEEE International Conference on Computer Vision, {ICCV}},
  2019.

\bibitem{maskrcnn}
K.~He, G.~Gkioxari, P.~Doll{\'{a}}r, and R.~B. Girshick.
\newblock Mask {R-CNN}.
\newblock In {\em {IEEE} International Conference on Computer Vision, {ICCV}},
  2017.

\bibitem{he2016deep}
K.~He, X.~Zhang, S.~Ren, and J.~Sun.
\newblock Deep residual learning for image recognition.
\newblock In {\em IEEE Conference on Computer Vision and Pattern Recognition,
  {CVPR}}, 2016.

\bibitem{he2019bag}
T.~He, Z.~Zhang, H.~Zhang, Z.~Zhang, J.~Xie, and M.~Li.
\newblock Bag of tricks for image classification with convolutional neural
  networks.
\newblock In {\em IEEE Conference on Computer Vision and Pattern Recognition,
  {CVPR}}, 2019.

\bibitem{hebb}
D.~O. Hebb.
\newblock {\em The organization of behavior: a neuropsychological theory}.
\newblock J. Wiley; Chapman \& Hall, 1949.

\bibitem{hinton2012practical}
G.~E. Hinton.
\newblock A practical guide to training restricted boltzmann machines.
\newblock In {\em Neural networks: Tricks of the trade}, pages 599--619.
  Springer, 2012.

\bibitem{dbn}
G.~E. Hinton, S.~Osindero, and Y.~W. Teh.
\newblock A fast learning algorithm for deep belief nets.
\newblock {\em Neural Computation}, 18(7):1527--1554, 2006.

\bibitem{hu2018squeeze}
J.~Hu, L.~Shen, and G.~Sun.
\newblock Squeeze-and-excitation networks.
\newblock In {\em IEEE Conference on Computer Vision and Pattern Recognition,
  {CVPR}}, 2018.

\bibitem{huang2019gpipe}
Y.~Huang, Y.~Cheng, A.~Bapna, O.~Firat, D.~Chen, M.~Chen, H.~Lee, J.~Ngiam,
  Q.~V. Le, Y.~Wu, et~al.
\newblock Gpipe: Efficient training of giant neural networks using pipeline
  parallelism.
\newblock In {\em Advances in Neural Information Processing Systems,
  {NeurIPS}}, pages 103--112, 2019.

\bibitem{synthetic}
M.~Jaderberg, W.~M. Czarnecki, S.~Osindero, O.~Vinyals, A.~Graves, D.~Silver,
  and K.~Kavukcuoglu.
\newblock Decoupled neural interfaces using synthetic gradients.
\newblock In {\em 34th International Conference on Machine Learning, {ICML}},
  2017.

\bibitem{kolesnikov2019revisiting}
A.~Kolesnikov, X.~Zhai, and L.~Beyer.
\newblock Revisiting self-supervised visual representation learning.
\newblock In {\em IEEE Conference on Computer Vision and Pattern Recognition,
  {CVPR}}, 2019.

\bibitem{weirdtrick}
A.~Krizhevsky.
\newblock One weird trick for parallelizing convolutional neural networks.
\newblock {\em CoRR}, abs/1404.5997, 2014.

\bibitem{alexnet}
A.~Krizhevsky, I.~Sutskever, and G.~E. Hinton.
\newblock Imagenet classification with deep convolutional neural networks.
\newblock In {\em Advances in Neural Information Processing Systems, {NIPS}},
  2012.

\bibitem{li2019selective}
X.~Li, W.~Wang, X.~Hu, and J.~Yang.
\newblock Selective kernel networks.
\newblock In {\em IEEE Conference on Computer Vision and Pattern Recognition,
  {CVPR}}, 2019.

\bibitem{mscoco}
T.~Lin, M.~Maire, S.~J. Belongie, J.~Hays, P.~Perona, D.~Ramanan,
  P.~Doll{\'{a}}r, and C.~L. Zitnick.
\newblock Microsoft {COCO:} common objects in context.
\newblock In D.~J. Fleet, T.~Pajdla, B.~Schiele, and T.~Tuytelaars, editors,
  {\em 13th European Conference on Computer Vision, {ECCV}}, 2014.

\bibitem{loshchilov2016sgdr}
I.~Loshchilov and F.~Hutter.
\newblock Sgdr: Stochastic gradient descent with warm restarts.
\newblock {\em arXiv preprint arXiv:1608.03983}, 2016.

\bibitem{e2e2e}
S.~L{\"{o}}we, P.~O'Connor, and B.~S. Veeling.
\newblock Putting an end to end-to-end: Gradient-isolated learning of
  representations.
\newblock In {\em Advances in Neural Information Processing Systems,
  {NeurIPS}}, 2019.

\bibitem{ma2018shufflenet}
N.~Ma, X.~Zhang, H.-T. Zheng, and J.~Sun.
\newblock Shufflenet v2: Practical guidelines for efficient cnn architecture
  design.
\newblock In {\em 15th European Conference on Computer Vision, {ECCV}}, 2018.

\bibitem{maaten2008visualizing}
L.~v.~d. Maaten and G.~Hinton.
\newblock Visualizing data using t-sne.
\newblock {\em Journal of machine learning research}, 9(Nov):2579--2605, 2008.

\bibitem{metaunsup}
L.~Metz, N.~Maheswaranathan, B.~Cheung, and J.~Sohl{-}Dickstein.
\newblock Meta-learning update rules for unsupervised representation learning.
\newblock In {\em 7th International Conference on Learning Representations,
  {ICLR}}, 2019.

\bibitem{backpropamine}
T.~Miconi, A.~Rawal, J.~Clune, and K.~O. Stanley.
\newblock Backpropamine: training self-modifying neural networks with
  differentiable neuromodulated plasticity.
\newblock In {\em 7th International Conference on Learning Representations,
  {ICLR}}, 2019.

\bibitem{diffplasticity}
T.~Miconi, K.~O. Stanley, and J.~Clune.
\newblock Differentiable plasticity: training plastic neural networks with
  backpropagation.
\newblock In J.~G. Dy and A.~Krause, editors, {\em 35th International
  Conference on Machine Learning, {ICML}}, 2018.

\bibitem{pirl}
I.~Misra and L.~van~der Maaten.
\newblock Self-supervised learning of pretext-invariant representations.
\newblock {\em CoRR}, abs/1912.01991, 2019.

\bibitem{pipedream}
D.~Narayanan, A.~Harlap, A.~Phanishayee, V.~Seshadri, N.~R. Devanur, G.~R.
  Ganger, P.~B. Gibbons, and M.~Zaharia.
\newblock Pipedream: Generalized pipeline parallelism for dnn training.
\newblock In {\em 27th ACM Symposium on Operating Systems Principles, {SOSP}},
  2019.

\bibitem{dfa}
A.~N{\o}kland.
\newblock Direct feedback alignment provides learning in deep neural networks.
\newblock In D.~D. Lee, M.~Sugiyama, U.~von Luxburg, I.~Guyon, and R.~Garnett,
  editors, {\em Advances in Neural Information Processing Systems 29: Annual
  Conference on Neural Information Processing Systems, {NeurIPS}}, 2016.

\bibitem{jigsaw}
M.~Noroozi and P.~Favaro.
\newblock Unsupervised learning of visual representations by solving jigsaw
  puzzles.
\newblock In B.~Leibe, J.~Matas, N.~Sebe, and M.~Welling, editors, {\em 14th
  European Conference on Computer Vision - {ECCV}}, 2016.

\bibitem{counting}
M.~Noroozi, H.~Pirsiavash, and P.~Favaro.
\newblock Representation learning by learning to count.
\newblock In {\em {IEEE} International Conference on Computer Vision, {ICCV}},
  2017.

\bibitem{oja}
E.~Oja.
\newblock Simplified neuron model as a principal component analyzer.
\newblock {\em Journal of mathematical biology}, 15(3):267--273, 1982.

\bibitem{peng2018megdet}
C.~Peng, T.~Xiao, Z.~Li, Y.~Jiang, X.~Zhang, K.~Jia, G.~Yu, and J.~Sun.
\newblock Megdet: A large mini-batch object detector.
\newblock In {\em IEEE Conference on Computer Vision and Pattern Recognition,
  {CVPR}}, 2018.

\bibitem{brainscore}
M.~Schrimpf, J.~Kubilius, H.~Hong, N.~J. Majaj, R.~Rajalingham, E.~B. Issa,
  K.~Kar, P.~Bashivan, J.~Prescott-Roy, K.~Schmidt, D.~L.~K. Yamins, and J.~J.
  DiCarlo.
\newblock Brain-score: Which artificial neural network for object recognition
  is most brain-like?
\newblock {\em bioRxiv}, 10.1101/407007, 2018.

\bibitem{dataparallel}
C.~J. Shallue, J.~Lee, J.~Antognini, J.~Sohl-Dickstein, R.~Frostig, and G.~E.
  Dahl.
\newblock Measuring the effects of data parallelismon neural network training.
\newblock {\em Journal of Machine Learning Research}, 20, 2019.

\bibitem{bm}
P.~Smolensky.
\newblock Information processing in dynamical systems: Foundations of harmony
  theory.
\newblock Technical report, Colorado Univ at Boulder Dept of Computer Science,
  1986.

\bibitem{cmc}
Y.~Tian, D.~Krishnan, and P.~Isola.
\newblock Contrastive multiview coding.
\newblock {\em CoRR}, abs/1906.05849, 2019.

\bibitem{cpc}
A.~van~den Oord, Y.~Li, and O.~Vinyals.
\newblock Representation learning with contrastive predictive coding.
\newblock {\em CoRR}, abs/1807.03748, 2018.

\bibitem{wu2018group}
Y.~Wu and K.~He.
\newblock Group normalization.
\newblock In {\em 15th European Conference on Computer Vision, {ECCV}}, 2018.

\bibitem{learn2remember}
Y.~Xiong, M.~Ren, and R.~Urtasun.
\newblock Learning to remember from a multi-task teacher.
\newblock {\em CoRR}, abs/1910.04650, 2019.

\bibitem{you2017large}
Y.~You, I.~Gitman, and B.~Ginsburg.
\newblock Large batch training of convolutional networks.
\newblock {\em arXiv preprint arXiv:1708.03888}, 2017.

\bibitem{colorization}
R.~Zhang, P.~Isola, and A.~A. Efros.
\newblock Colorful image colorization.
\newblock In B.~Leibe, J.~Matas, N.~Sebe, and M.~Welling, editors, {\em 14th
  European Conference on Computer Vision, {ECCV}}, 2016.

\bibitem{contrastivebrain}
C.~Zhuang, S.~Yan, A.~Nayebi, M.~Schrimpf, M.~C. Frank, J.~J. DiCarlo, and
  D.~L.~K. Yamins.
\newblock Unsupervised neural network models of the ventral visual stream.
\newblock {\em bioRxiv}, 10.1101/2020.06.16.155556, 2020.

\bibitem{localagg}
C.~Zhuang, A.~L. Zhai, and D.~Yamins.
\newblock Local aggregation for unsupervised learning of visual embeddings.
\newblock In {\em {IEEE/CVF} International Conference on Computer Vision,
  {ICCV}}, 2019.

\end{thebibliography}


\begin{thebibliography}{1}

\bibitem{chen2020improved}
X.~Chen, H.~Fan, R.~Girshick, and K.~He.
\newblock Improved baselines with momentum contrastive learning.
\newblock {\em arXiv preprint arXiv:2003.04297}, 2020.

\bibitem{maaten2008visualizing}
L.~v.~d. Maaten and G.~Hinton.
\newblock Visualizing data using t-sne.
\newblock {\em Journal of machine learning research}, 9(Nov):2579--2605, 2008.

\end{thebibliography}
\bibliographystyle{abbrv}
}
\clearpage
\appendix
\section{Training Curves}
\begin{figure}[htbp]
\centering
\begin{minipage}[t]{0.48\linewidth}
\includegraphics[width=1\linewidth, trim={0.3cm 0 1cm 0}, clip]{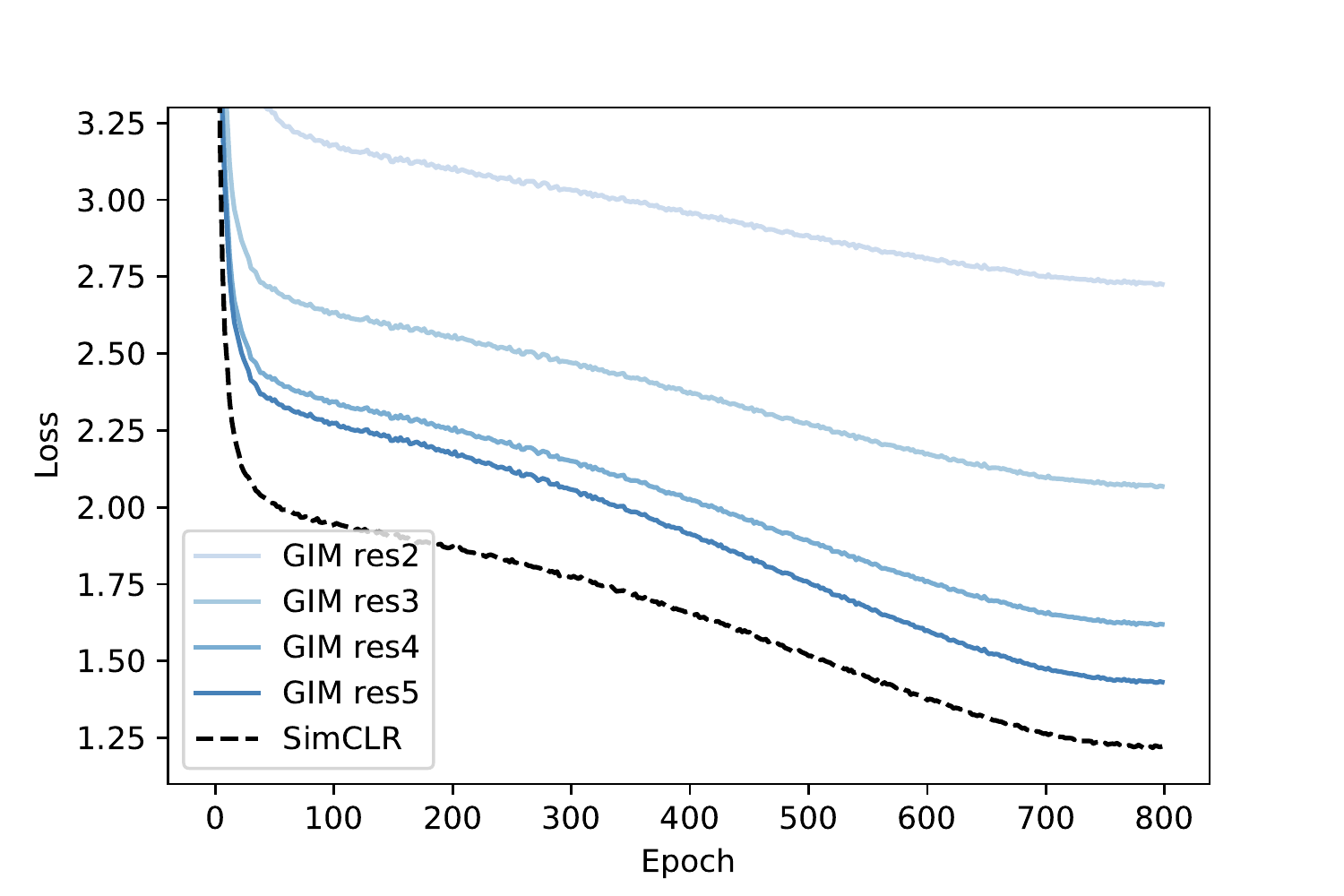}
\end{minipage}
\hfill
\begin{minipage}[t]{0.48\linewidth}
\includegraphics[width=1\linewidth, trim={0.3cm 0 1cm 0}, clip]{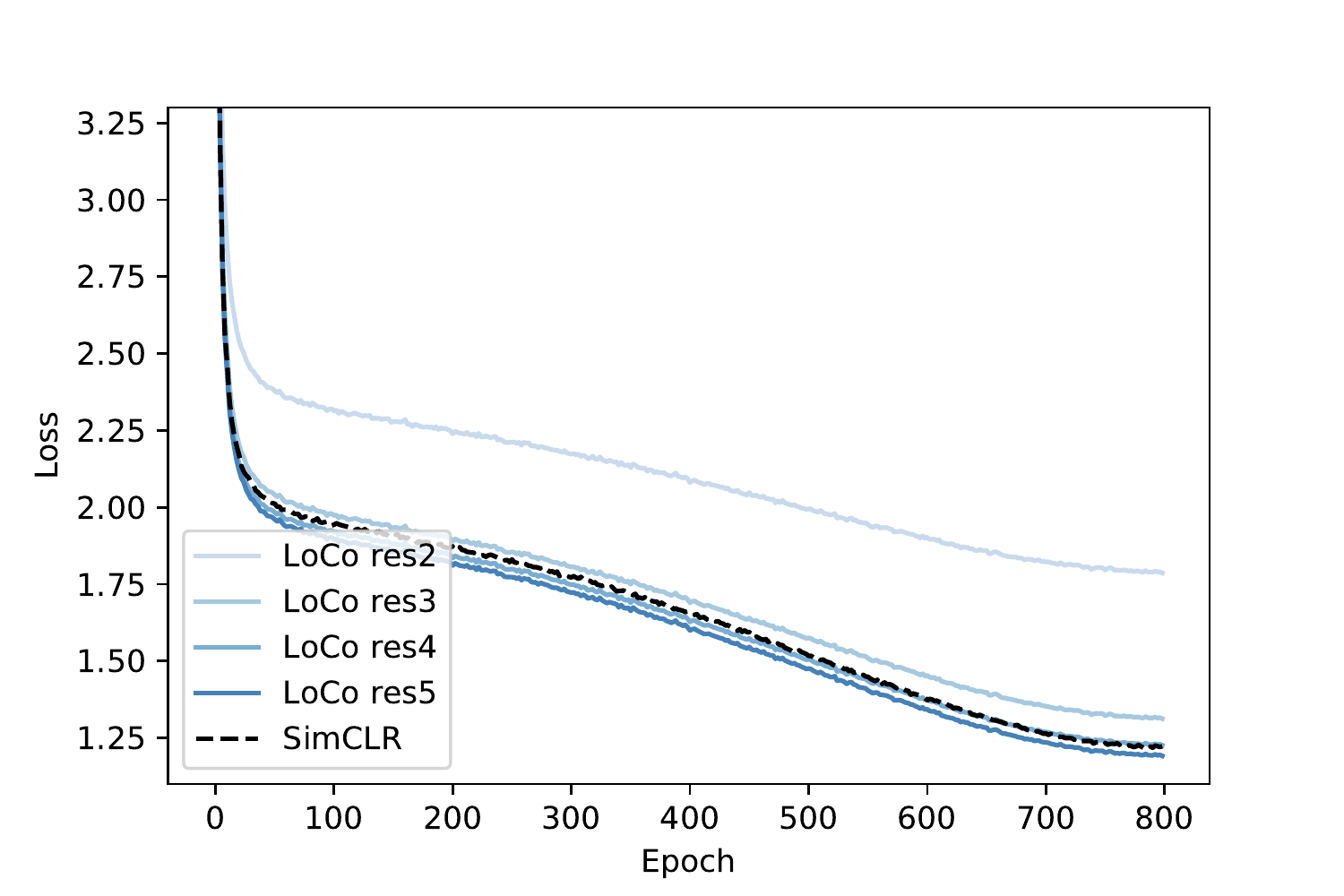}	
\end{minipage}
\caption{Training loss curves for SimCLR, GIM and \ours{}}
\label{fig:loss_curve}
\end{figure}

We provide training loss curves of SimCLR, GIM and \ours{} for a better understanding of the
performance gap between them. Contrastive losses computed using outputs from the full ResNet-50
encoder are shown in Fig.~\ref{fig:loss_curve}. For GIM and \ours{}, losses from other decoders,
including {\em res2}, {\em res3}, {\em res4}, are also provided. As we can see in
Fig.~\ref{fig:loss_curve}, the losses of different decoders in \ours{}  closely match the loss of
the decoder in SimCLR during training, with the exception of {\em res2}, while for GIM this is not
the case.

\section{Architecture of Progressive ResNet-50}

In this section we show the block structure of each stage in Progressive ResNet-50 in
Table~\ref{tab:presnet_arch}. The block structure of ResNet-50 is also shown here for reference. We
downsample the feature map size progressively using bilinear interpolation, and  use basic blocks to
reduce the memory footprint of earlier stages, and group convolution to reduce the computation cost
of later stages to get a model with more balanced computation and memory footprint at each stage. We
use this model to show the great potential of \ours{} in terms of both memory saving and computation
for model parallelism. As it is designed to have 15\textasciitilde 20 memory footprint and
computation cost per stage, the peak memory usage will be significantly reduced in local learning,
and no worker that handles a stage of the encoder will become a computation bottleneck in model
parallelism.

\begin{table}[htbp]

\centering
\begin{tabular}{|c|c|c|c|c|}
\hline
\multirow{2}*{layer}     &  \multicolumn{2}{c|}{PResNet-50} & \multicolumn{2}{c|}{ResNet-50} \\ 
\cline{2-5} & output size & block structure & output size & block structure\\
\hline
conv1  & 112$\times$112    & 7x7, 32, stride 2 & 112$\times$112 &  7$\times$7, 64, stride 2  \\ 
\hline
\multirow{2}{*}{res2\_x} & \multirow{2}{*}{56$\times$56} & 3$\times$3 max pool, stride 2 & \multirow{2}{*}{56$\times$56} & 3$\times$3 max pool, stride 2\\ 
\cline{3-3} \cline{5-5} &                        
& \begin{math} \left[ \begin{tabular}[c]{@{}c@{}}3$\times$3, 56\\ 3$\times$3, 56\end{tabular}  \right ] \end{math}$\times$3  & ~ & \begin{math} \left[ \begin{tabular}[c]{@{}c@{}}1$\times$1, 64\\ 3$\times$3, 64\\ 1$\times$1, 256 \end{tabular}  \right ] \end{math}$\times$3   \\ 
\hline
res3\_x                  
& 36$\times$36 
& \begin{math} \left[ \begin{tabular}[c]{@{}c@{}}3$\times$3, 96\\ 3$\times$3, 96 \end{tabular} \right ] \end{math}$\times$3 
& 28$\times$28 
&  \begin{math} \left[  \begin{tabular}[c]{@{}c@{}}1$\times$1, 128\\  3$\times$3, 128\\ 1$\times$1, 512 \end{tabular}  \right ] \end{math}$\times$4  \\ 
\hline
res4\_x                  & 24$\times$24                  
& \begin{math} \left[ \begin{tabular}[c]{@{}c@{}}1$\times$1, 144\\ 3$\times$3, 144\\ 1$\times$1, 576 \end{tabular} \right ] \end{math}$\times$3 
& 14$\times$14
& \begin{math} \left[ \begin{tabular}[c]{@{}c@{}}1$\times$1, 256\\  3$\times$3, 256\\ 1$\times$1, 1024 \end{tabular} \right ] \end{math}$\times$6 \\ 
\hline
res5\_x                  & 16$\times$16                    
& \begin{tabular}[c]{@{}c@{}} \begin{math} \left[ \begin{tabular}[c]{@{}c@{}}1$\times$1, 256, 2 grps\\ 3$\times$3, 256, 2 grps\\ 1$\times$1, 1024 \end{tabular} \right ] \end{math}$\times$1 \\ \begin{math} \left[ \begin{tabular}[c]{@{}c@{}}1$\times$1, 256, 2 grps\\ 3$\times$3, 256, 2 grps\\ 1$\times$1, 1024, 2 grps \end{tabular} \right ] \end{math}$\times$2\end{tabular}
& 7$\times$7
& \begin{math} \left[ \begin{tabular}[c]{@{}c@{}}1$\times$1, 512\\ 3$\times$3, 512\\ 1$\times$1, 2048 \end{tabular} \right ] \end{math}$\times$3 \\ \hline
res6\_x                  & 12$\times$12                    
& \begin{tabular}[c]{@{}c@{}} \begin{math} \left[ \begin{tabular}[c]{@{}c@{}}1$\times$1, 512, 16 grps\\ 3$\times$3, 512, 16 grps\\ 1$\times$1, 2048 \end{tabular} \right ] \end{math}$\times$1\\ \begin{math} \left[ \begin{tabular}[c]{@{}c@{}}1$\times$1, 512, 16 grps\\ 3$\times$3, 512, 16 grps\\ 1$\times$1, 2048, 16 grps \end{tabular} \right ] \end{math}$\times$2 \end{tabular} & - & -\\ \hline
res7\_x                  & 8$\times$8                    
& \begin{tabular}[c]{@{}c@{}} \begin{math} \left[ \begin{tabular}[c]{@{}c@{}}1$\times$1, 1024, 128 grps\\ 3$\times$3, 1024, 128 grps\\ 1$\times$1, 4096 \end{tabular} \right ] \end{math} $\times$1 \\ \begin{math} \left[ \begin{tabular}[c]{@{}c@{}}1$\times$1, 1024, 128 grps\\ 3$\times$3, 1024, 128 grps\\ 1$\times$1, 4096, 128 grps \end{tabular} \right ] \end{math} $\times$2 \end{tabular} & -  & -\\ 
\hline
                          & 1$\times$1                    & average pool, 1000-d fc & 1$\times$1 & average pool, 1000-d fc                                                                                                                                                                                                                                                                       \\ \hline
\end{tabular}
\vspace{0.1in}
\caption{Architectural details of Progressive ResNet-50 and ResNet-50. Output sizes for both models are specified individually}
\label{tab:presnet_arch}
\end{table}

\section{Representation visualization}

\begin{figure}[htbp]
\centering
\begin{minipage}[t]{0.31\linewidth}
\includegraphics[width=1\linewidth, trim={1.9cm 0 1.5cm 0}, clip]{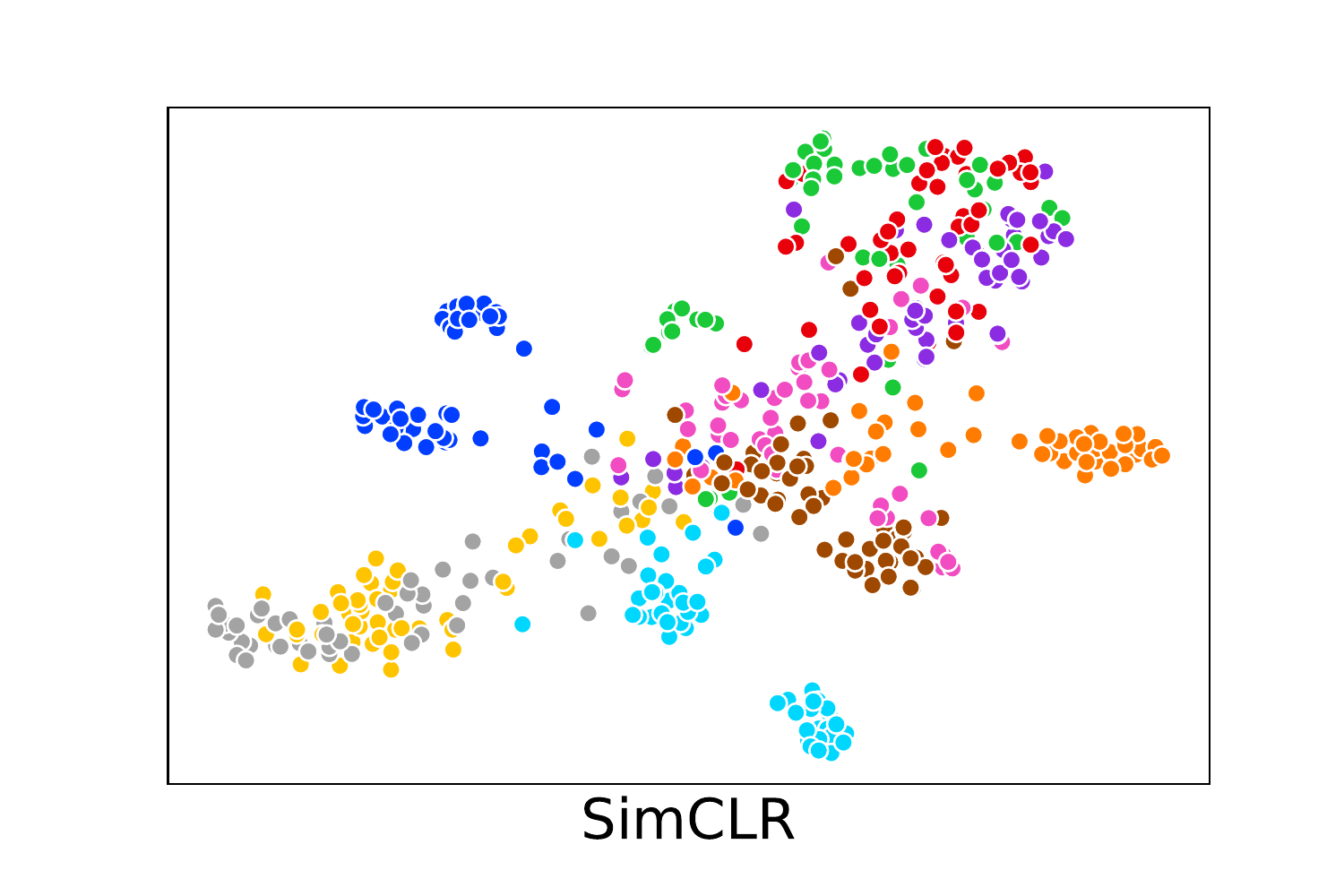}
\end{minipage}
\hfill
\begin{minipage}[t]{0.31\linewidth}
\includegraphics[width=1\linewidth, trim={1.9cm 0 1.5cm 0}, clip]{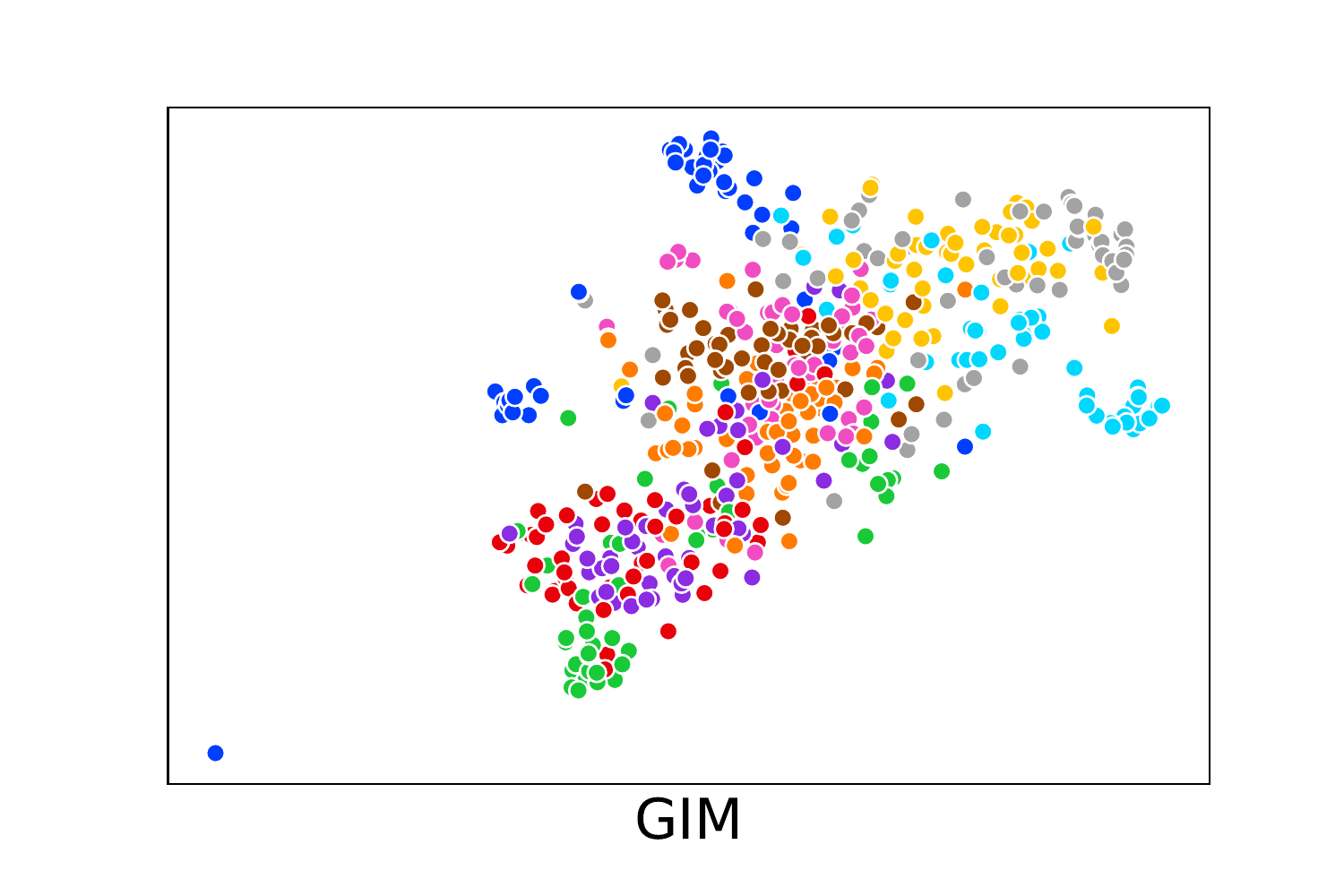}
\end{minipage}
\hfill
\begin{minipage}[t]{0.31\linewidth}
\includegraphics[width=1\linewidth, trim={1.9cm 0 1.5cm 0}, clip]{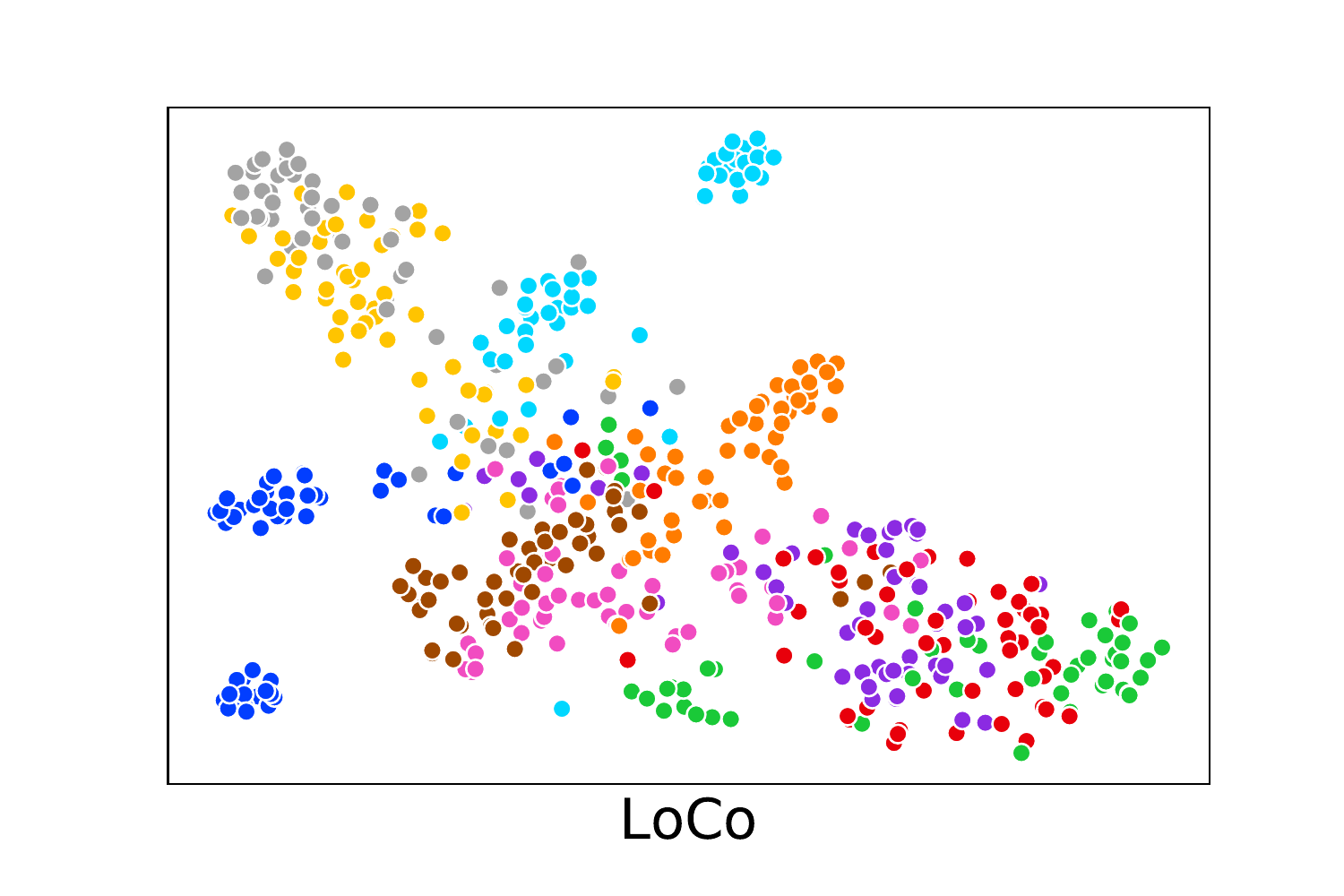}	
\end{minipage}
\caption{t-SNE visualization results for SimCLR, GIM and \ours{}}
\label{fig:tsne_vis}
\end{figure}

In this section we show some visualization results of the learned representation
of SimCLR, GIM and
\ours{}. We subsample images belonging to the first 10 classes of ImageNet-1K from
the validation set (500 images in total) and use t-SNE~\cite{maaten2008visualizing} to visualize the
4096-d vector representation from the PResNet-50 encoder. The results are shown in
Fig~\ref{fig:tsne_vis}. We can see \ours{} learns image
embedding vectors that can form more compact clusters compared to GIM.

\clearpage

\section{Qualitative results for downstream tasks}

\begin{figure}[htbp]
\centering
\includegraphics[width=1.0\linewidth]{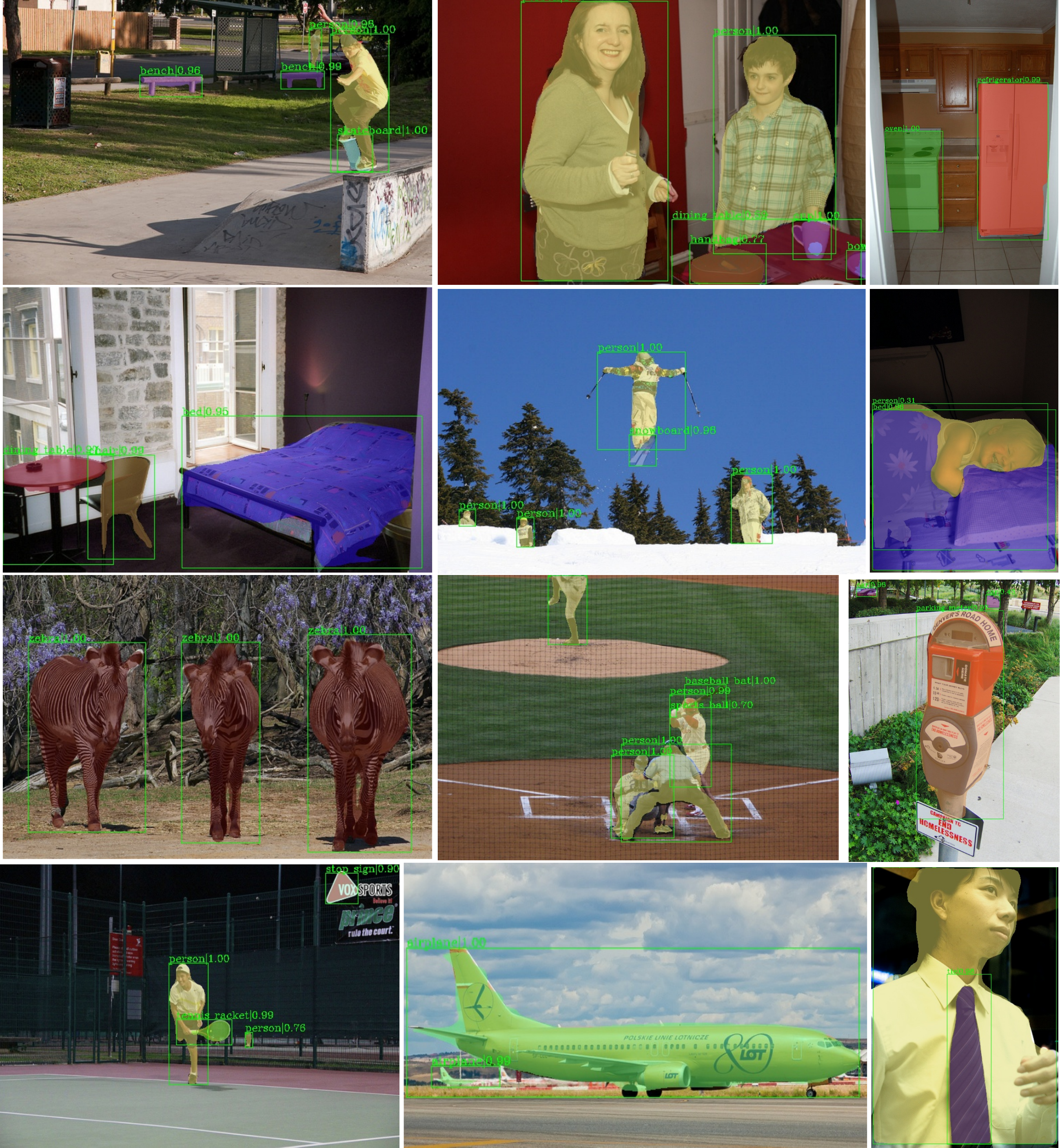}	
\caption{Qualitative results on COCO-10K, \ours{} trained on 800 epochs with ResNet-50 is used to
initialize the Mask R-CNN model}
\label{fig:coco_vis}
\end{figure}

Last, we show qualitative results of detection and instance segmentation tasks on COCO in Fig.~\ref{fig:coco_vis}.

\newpage
\end{document}